\documentclass[twoside,11pt]{article}

\usepackage[preprint,abbrvbib,nohyperref]{jmlr2e}
\usepackage{amsmath}
\usepackage{rotating}

\begin{document}

\title{Efficient computation of rankings from pairwise comparisons}

\author{\name M. E. J. Newman\\
       \addr Center for the Study of Complex Systems\\
       University of Michigan\\
       Ann Arbor, MI 48109, USA}

\maketitle

\begin{abstract}
We study the ranking of individuals, teams, or objects, based on pairwise comparisons between them, using the Bradley-Terry model.  Estimates of rankings within this model are commonly made using a simple iterative algorithm first introduced by Zermelo almost a century ago.  Here we describe an alternative and similarly simple iteration that provably returns identical results but does so much faster---over a hundred times faster in some cases.  We demonstrate this algorithm with applications to a range of example data sets and derive a number of results regarding its convergence.
\end{abstract}

\section{Introduction}
The problem of ranking a set of individuals, teams, or objects on the basis of a set of pairwise comparisons between them arises in many contexts, including competitions in sports, chess, and other games, paired comparison studies of consumer choice, and observational studies of dominance behaviors in animals and humans \citep{Zermelo29,BT52,DF76,David88,Cattelan12}.  If a group of chess players play games against one another, for example, how can we rank the players, from best to worst, based on the outcome of those games?  The outcomes may be contradictory or ambiguous---underdogs sometimes win and strong players sometimes lose---so we adopt a probabilistic model.  In the most common version we assign a numerical score~$s_i$ to each individual~$i$ and the probability~$p_{ij}$ of $i$ beating~$j$ is assumed to be some function of the difference in their scores $p_{ij}=f(s_i-s_j)$.  The most popular choice of functional form is the logistic function $f(s) = 1/(1+e^{-s})$, which gives
\begin{equation}
p_{ij} = {e^{s_i}\over e^{s_i}+e^{s_j}}.
\label{eq:btwin}
\end{equation}
This is the Bradley-Terry model, first introduced by \citet{Zermelo29} and heavily studied in the years since, particularly following its rediscovery by \citet{BT52}.  For convenience, one often introduces the shorthand $\pi_i = e^{s_i}$ so that $p_{ij}$ can also be written as
\begin{equation}
p_{ij}  = {\pi_i\over\pi_i+\pi_j},
\end{equation}
and we will do that here.  Zermelo (writing in German) referred to the non-negative parameters~$\pi_i$ as \textit{Spielst\"arken} or ``playing strengths,'' although they are elsewhere variously called worth parameters, skill parameters, merit parameters, ratings, or simply weights.  Following Zermelo, we will call them strengths.

Given the outcomes of a series of pairwise competitions between $N$ competitors the strengths can be estimated straightforwardly.  Commonly one makes a maximum-likelihood estimate.  Defining $w_{ij}$ to be the total number of times $i$ beats~$j$, or zero if $i$ and~$j$ never competed, it can be shown that the maximum-likelihood values of the strengths are given by a simple procedure: starting from any convenient initial values we iterate the equation
\begin{equation}
\pi_i' = {\sum_{j=1}^N w_{ij}\over\sum_{j=1}^N (w_{ij}+w_{ji})/(\pi_i+\pi_j)}
\label{eq:tired}
\end{equation}
until convergence is reached.  This algorithm was also first described by \citet{Zermelo29} and we will refer to it as Zermelo's algorithm.  An extraordinary number of papers have been written about it, its variants, its properties, and its applications.

Although widely used, however, Zermelo's algorithm is known to be slow to converge \citep{Dykstra56,Hunter04}.  In this paper we study the alternate iteration
\begin{equation}
\pi_i' = {\sum_{j=1}^Nw_{ij} \pi_j/(\pi_i+\pi_j)\over
              \sum_{j=1}^N w_{ji}/(\pi_i+\pi_j)}.
\label{eq:wired}
\end{equation}
We show that iteration of this equation solves the same problem and converges to the same solution as Zermelo's algorithm but does so significantly faster---over a hundred times faster in some cases.  Given that Eq.~\eqref{eq:wired} is also simple to implement we know of no reason not to favor it over Eq.~\eqref{eq:tired}.

In recent years a number of other authors have also considered alternative and potentially more efficient algorithms for ranking under the Bradley-Terry model.  One promising approach employs spectral methods that estimate rankings based on the properties of random walks on the network of directed interactions between individuals \citep{MG15,NOS17,APA18}.  Although they do not directly maximize likelihood under the Bradley-Terry model, these algorithms can be shown to converge closely to the maximum-likelihood solution.  Some versions are quite numerically efficient, though they can also be complex to implement.  Minorization-maximization (MM) algorithms, which optimize a minorizing proxy for the likelihood function, can also be applied to ranking problems.  For the Bradley-Terry model the appropriate MM algorithm turns out to be exactly equivalent to Zermelo's algorithm and hence offers no speed improvement \citep{Hunter04}, but techniques have been suggested for accelerating convergence \citep{VYZ19} and the MM formulation also provides an elegant route to developing algorithms for generalizations of the model.  Perhaps more directly competitive with our approach is one of the simplest of methods: one can fit the Bradley-Terry model using Newton's method applied to the derivative of the likelihood.  For small applications with only a few individuals or teams to be ranked this is typically the fastest approach, although for such small cases the difference may be moot.  As the number~$N$ of individuals gets larger, however, Newton's method becomes impractical because the time taken per iteration of the algorithm scales as~$N^3$, which quickly becomes prohibitive \citep{Hunter04}.  For Eq.~\eqref{eq:wired} the time per iteration scales only as the number of pairwise competitions, making this approach faster for larger applications.  Maximum-likelihood methods are not the only approach for fitting: one can also adopt Bayesian approaches, although these usually require Monte Carlo estimation and hence are not competitive in terms of speed \citep{DS73,CD12}.  Overall, the particular combination of simplicity and speed offered by Eq.~\eqref{eq:wired} makes it an attractive approach for practical applications.

The rest of this paper is organized as follows.  In Section~\ref{sec:bt} we describe and derive Zermelo's original ranking algorithm and the new algorithm proposed here.  In Section~\ref{sec:convergence} we prove that the new algorithm converges to the same unique maximum-likelihood solution as Zermelo's algorithm.  In Section~\ref{sec:general} we briefly discuss a larger family of ranking algorithms of which Zermelo's algorithm and our own are special cases.  Within this family, the algorithm of this paper is the fastest to converge and hence it is our primary focus.  In Sections~\ref{sec:prior} and~\ref{sec:ties} we describe two extensions of our approach, one to maximum a priori (MAP) estimation of rankings and the other to a (previously proposed) generalization of the Bradley-Terry model that allows for ties or draws in competitions.  In Section~\ref{sec:results} we apply our algorithms to a broad selection of example data, both real and synthetic, finding in every case that the algorithm of this paper is faster than Zermelo's, often by a wide margin.  In Section~\ref{sec:conclusions} we give our conclusions.  Some additional technical results are presented in appendices.

\section{Iterative algorithms and the Bradley-Terry model}
\label{sec:bt}
Consider a tournament where $N$ players or teams play games of some kind against one another in pairs.  We will initially assume that no ties or draws are allowed, so there is a clear winner and loser of every game.  The case where ties are allowed is treated separately in Section~\ref{sec:ties}.  We assume that the probability~$p_{ij}$ that player~$i$ beats player~$j$ obeys Eq.~\eqref{eq:btwin} and we consider the strengths~$\pi_i$ to be a measure of the skill of the players, higher values indicating better players.

Note that the win probabilities~$p_{ij}$ are invariant under multiplication of all the $\pi_i$ by any constant.  One can remove this ambiguity by imposing any convenient normalization condition.  Here we fix the geometric mean strength to be~1, which is equivalent to setting $\prod_i \pi_i = 1$.  This choice has the nice effect that the probability~$p_1$ of a player with strength~$\pi$ beating the average player with strength~1 is $p_1 = \pi/(\pi+1)$ and hence $\pi = p_1/(1-p_1)$.  Thus the strength parameter has a simple interpretation: it is the odds of beating the average player.

\subsection{Zermelo's algorithm}
Suppose the tournament consists of a total of~$M$ games between pairs of players and let $w_{ij}$ be the number of times that player~$i$ beats player~$j$.  Using these data we can make a maximum-likelihood estimate of the strengths~$\pi_i$ as follows.  The likelihood of the observed games given the strengths (represented by a matrix $W = [w_{ij}]$ and vector~$\pi = [\pi_i]$ respectively) is
\begin{equation}
P(W|\pi) = \prod_{ij} p_{ij}^{w_{ij}
       } = \prod_{ij} \biggl( {\pi_i\over\pi_i+\pi_j} \biggr)^{w_{ij}},
\label{eq:likelihood}
\end{equation}
so that the log-likelihood is
\begin{equation}
\log P(W|\pi) = \sum_{ij} w_{ij} \log {\pi_i\over\pi_i+\pi_j}
  = \sum_{ij} w_{ij} \log \pi_i - \sum_{ij} w_{ij} \log(\pi_i+\pi_j).
\label{eq:loglike}
\end{equation}
Differentiating with respect to~$\pi_i$ for any~$i$ and setting the result to zero we get
\begin{equation}
{1\over\pi_i} \sum_j w_{ij} - \sum_j {w_{ij} + w_{ji}\over\pi_i + \pi_j} = 0,
\label{eq:deriv}
\end{equation}
which can be rearranged to read
\begin{equation}
\pi_i = {\sum_j w_{ij}\over\sum_j (w_{ij}+w_{ji})/(\pi_i+\pi_j)}.
\label{eq:fixedpoint}
\end{equation}
In general this equation has no closed-form solution but it can be solved numerically by simple iteration: one picks a suitable set of non-negative starting values for the~$\pi_i$---random values are often used---and then computes new values~$\pi_i'$ according to
\begin{equation}
\pi_i' = {\sum_j w_{ij}\over\sum_j (w_{ij}+w_{ji})/(\pi_i+\pi_j)}.
\label{eq:zermelo}
\end{equation}
Iterating this process, it can be proved subject to certain conditions that we converge to the global maximum of the likelihood and hence obtain an estimate of the strengths~$\pi_i$ \citep{Zermelo29,Ford57,Hunter04}.  The values can then be sorted in order to give a ranking of the players, or simply used in their raw form as a kind of rating.  The iteration can be performed synchronously (all $\pi_i$ updated at the same time) or asynchronously ($\pi_i$~updated one by one in cyclic fashion), but it is generally believed that asynchronous updating is more efficient, since the update of any individual~$\pi_i$ benefits from the improved estimates of previously updated ones.  In this paper we use asynchronous updates.  Our convergence results in Sections~\ref{sec:convergence} and~\ref{sec:general} are also for the asynchronous case.

This iterative algorithm, first described by \citet{Zermelo29}, is the standard method for calculating rankings within the Bradley-Terry model and has seen numerous applications over the years in a wide variety of contexts.

\subsection{An alternative algorithm}
\label{sec:new}
In a sense, computing maximum-likelihood estimates for the Bradley-Terry model is a straightforward problem.  As discussed in Section~\ref{sec:convergence}, the likelihood is concave and many standard convex optimization methods can be applied.  Speed, however, is of the essence in practical applications of the model, so considerable effort has been exerted in recent years to find solution methods faster than Zermelo's algorithm \citep{MG15,NOS17,APA18,VYZ19}.  Some of these are quite complex, but here we propose a very simple approach that also turns out to be highly efficient.

Grouping the terms slightly differently, Eq.~\eqref{eq:deriv} can be rewritten as
\begin{equation}
{1\over\pi_i} \sum_j w_{ij} {\pi_j\over\pi_i+\pi_j}
  - \sum_j {w_{ji}\over\pi_i+\pi_j} = 0,
\label{eq:newderiv}
\end{equation}
which can be rearranged as
\begin{equation}
\pi_i = {\sum_j w_{ij} \pi_j/(\pi_i+\pi_j)\over
            \sum_j w_{ji}/(\pi_i+\pi_j)}.
\label{eq:fixedpoint2}
\end{equation}
This suggests a different iterative algorithm for the Bradley-Terry model.  Again we choose suitable starting values (for instance at random), then we iterate the form
\begin{equation}
\pi_i' = {\sum_j w_{ij} \pi_j/(\pi_i+\pi_j)\over
            \sum_j w_{ji}/(\pi_i+\pi_j)}
\label{eq:new}
\end{equation}
to convergence.  In Section~\ref{sec:convergence} we prove that, like Zermelo's algorithm, this process converges to the global maximum of the likelihood.

One nice feature of this algorithm is that it is transparent from Eq.~\eqref{eq:new} that $\pi_i=0$ for any individual who loses all their games and $\pi_i=\infty$ for any individual who wins all their games.  Furthermore, the iteration converges to these values in a single step.  The same values are also returned by the standard Zermelo algorithm, but it is less obvious from Eq.~\eqref{eq:zermelo} that this is true---it is some work to demonstrate the result for the player who wins every game and moreover it takes the Zermelo algorithm an infinite number of iterations to reach the correct value instead of just one iteration.  This is a special case of the more general finding, which we explore in this paper, that~\eqref{eq:new} converges faster than Zermelo's algorithm.

\section{Convergence}
\label{sec:convergence}
In this section we prove that the iteration of Eq.~\eqref{eq:new} converges to the global maximum of the likelihood, Eq.~\eqref{eq:likelihood}, from any starting point, whenever a maximum exists.

Zermelo proved that the likelihood has only one stationary point for $\pi_i\ge0$, corresponding to the global maximum, provided the $\pi_i$ are normalized as discussed in Section~\ref{sec:bt} and the directed network of interactions (the network with adjacency matrix~$w_{ij}$) is strongly connected, i.e.,~there is a directed path through the network from every individual to every other \citep{Zermelo29,Ford57}.  If the network is not strongly connected then there are no stationary points and there is no maximum of the likelihood, and hence our problem has no solution.  For the moment we will assume, as other authors have done, that the network is strongly connected and hence that there is a maximum of the likelihood, although we show how to relax this requirement in Section~\ref{sec:prior}.

Since any fixed point of the iteration of Eq.~\eqref{eq:new} corresponds to a stationary point of the likelihood, and since the iteration generates non-negative values of~$\pi_i$ only (given non-negative initial values), it follows that if the iteration converges to a fixed point that point must be the global maximum.  To prove that it converges to a fixed point it suffices to demonstrate that the value of the log-likelihood always increases upon application of Eq.~\eqref{eq:new} unless a fixed point has been reached, since the log-likelihood cannot increase without bound, being bounded above by the maximum.

We consider the asynchronous version of the iteration of Eq.~\eqref{eq:new} in which a single~$\pi_i$ is updated at each step, all others~$\pi_j$ remaining the same.  The $\pi_i$ are updated in order until all $N$ have been updated.  Consider the step on which a particular~$\pi_i$ is updated.  We define a function~$f(\pi_i)$ equal to the sum of the terms in the log-likelihood, Eq.~\eqref{eq:loglike}, that depend on~$\pi_i$:
\begin{equation}
f(\pi_i) = \sum_j w_{ij} \log {\pi_i\over\pi_i+\pi_j}
               - \sum_j w_{ji} \log(\pi_i+\pi_j).
\label{eq:f}
\end{equation}
Noting that $\log x \le x - 1$ for all real $x>0$ and making the substitution $x\to x/y$, we derive the useful inequality
\begin{equation}
\log x \le \log y + {x\over y} - 1
\label{eq:ineq1}
\end{equation}
for all $x,y>0$, or equivalently
\begin{equation}
\log y \ge \log x - {x\over y} + 1,
\label{eq:ineq2}
\end{equation}
with the exact equality holding if and only if $x=y$.  This implies for any~$\pi_i$ and~$\pi_i'$ that
\begin{equation}
\log {\pi_i'\over\pi_i'+\pi_j} \ge \log {\pi_i\over\pi_i+\pi_j}
  - {\pi_i/(\pi_i+\pi_j)\over\pi_i'/(\pi_i'+\pi_j)} + 1
  = \log {\pi_i\over\pi_i+\pi_j}
     + {(\pi_i'-\pi_i)/\pi_i'\over(\pi_i+\pi_j)/\pi_j},
\label{eq:piineq1}
\end{equation}
and \eqref{eq:ineq1} implies that
\begin{equation}
\log(\pi_i'+\pi_j) \le \log(\pi_i+\pi_j) + {\pi_i'+\pi_j\over\pi_i+\pi_j} - 1
  = \log(\pi_i+\pi_j) + {\pi_i'-\pi_i\over\pi_i+\pi_j}.
\label{eq:piineq2}
\end{equation}
Evaluating Eq.~\eqref{eq:f} at the point~$\pi_i'$ defined by Eq.~\eqref{eq:new} and applying these two inequalities, we find that
\begin{align}
f(\pi_i')
  &= \sum_j w_{ij} \log {\pi_i'\over\pi_i'+\pi_j}
               - \sum_j w_{ji} \log(\pi_i'+\pi_j) \nonumber\\
  &\ge \sum_j w_{ij} \log {\pi_i\over\pi_i+\pi_j}
     + {\pi_i'-\pi_i\over\pi_i'}
       \sum_j w_{ij} {\pi_j\over\pi_i+\pi_j}
     \nonumber\\
  &\hspace{8em}{} - \sum_j w_{ji} \log (\pi_i+\pi_j)
   - (\pi_i'-\pi_i) \sum_j {w_{ji}\over\pi_i+\pi_j}
   \nonumber\\
  &= \sum_j w_{ij} \log {\pi_i\over\pi_i+\pi_j}
     - \sum_j w_{ji} \log (\pi_i+\pi_j)
     \nonumber\\
  &\hspace{8em}{} + (\pi_i'-\pi_i) \biggl[ {1\over\pi_i'}
   \sum_j w_{ij} {\pi_j\over\pi_i+\pi_j}
   -  \sum_j {w_{ji}\over\pi_i+\pi_j} \biggr] \nonumber\\
  &= f(\pi_i),
\end{align}
where we have used Eq.~\eqref{eq:f} again, the term inside the square brackets vanishes because of~\eqref{eq:new}, and the exact equality applies if and only if $\pi_i'=\pi_i$.

Thus $f(\pi_i)$ always increases upon application of Eq.~\eqref{eq:new} and hence so also does the log-likelihood, unless $\pi_i'=\pi_i$, in which case the log-likelihood remains the same but could still increase when one of the other $\pi_i$ is updated.  Only if $\pi_i'=\pi_i$ for all~$i$ does the log-likelihood not increase at all, but if this occurs then by definition we have reached a fixed point of the iteration, and hence we have reached the global maximum.  This now guarantees convergence of the iterative algorithm of Eq.~\eqref{eq:new} to the global likelihood maximum.

\section{Other iterative algorithms}
\label{sec:general}
Given the existence of two different iterations, Eqs.~\eqref{eq:zermelo} and~\eqref{eq:new}, that both converge to the same maximum-likelihood estimate, one might wonder whether there exist any others.  In fact, it turns out there is an entire one-parameter family of iterations that includes \eqref{eq:zermelo} and~\eqref{eq:new} as special cases, and all of them converge to the same solution.  Of these iterations, Eq.~\eqref{eq:new} converges most rapidly and hence is our primary focus in this paper, but for the interested reader we discuss the full family briefly in this section.

For any $\alpha$ we can rewrite Eq.~\eqref{eq:newderiv} in the form
\begin{equation}
{1\over\pi_i} \sum_j w_{ij} {\alpha\pi_i+\pi_j\over\pi_i+\pi_j}
  - \sum_j {\alpha w_{ij} + w_{ji}\over\pi_i+\pi_j} = 0,
\label{eq:genzero}
\end{equation}
which we can solve by iterating
\begin{equation}
\pi_i' = {\sum_j w_{ij} (\alpha\pi_i+\pi_j)/(\pi_i+\pi_j)\over
            \sum_j (\alpha w_{ij}+w_{ji})/(\pi_i+\pi_j)}
\label{eq:general}
\end{equation}
until convergence is achieved.  When $\alpha=1$ this procedure is equivalent to Zermelo's algorithm, Eq.~\eqref{eq:zermelo}.  When $\alpha=0$ it is equivalent to the algorithm presented in this paper, Eq.~\eqref{eq:new}.  For negative $\alpha$ the iteration does not generate positive values of~$\pi_i$ in general and hence is invalid, but for zero or positive~$\alpha$ it gives a whole range of algorithms, all of which converge to the same maximum-likelihood solution as Zermelo's algorithm.  When $0\le\alpha\le1$, convergence can be proved using a straightforward generalization of the method of Section~\ref{sec:convergence} as follows.

Since the log-likelihood has only a single stationary point corresponding to the global maximum, and since any fixed point of~\eqref{eq:general} is a solution of~\eqref{eq:genzero} and hence corresponds to a stationary point of the log-likelihood, it follows that if~\eqref{eq:general} converges to a fixed point at all it must converge to the global likelihood maximum.  To show that it converges to a fixed point it suffices, as previously, to show that the log-likelihood always increases upon application of~\eqref{eq:general} unless a fixed point has been reached.  To do this we rewrite Eq.~\eqref{eq:f} for the terms in the log-likelihood that depend on~$\pi_i$ as
\begin{equation}
f(\pi_i) = \alpha \sum_j w_{ij} \log \pi_i
  + (1-\alpha) \sum_j w_{ij} \log {\pi_i\over\pi_i+\pi_j}
  - \sum_j (\alpha w_{ij}+w_{ji}) \log (\pi_i+\pi_j).
\label{eq:fgeneral}
\end{equation}
For any $\pi_i,\pi_i'$ the inequality~\eqref{eq:ineq2} implies that
\begin{equation}
\log \pi_i' \ge \log \pi_i - {\pi_i\over\pi_i'} + 1
  = \log \pi_i + {\pi_i'-\pi_i\over\pi_i'},
\end{equation}
with the exact equality applying if and only if $\pi_i'=\pi_i$.  Evaluating Eq.~\eqref{eq:fgeneral} at the point~$\pi_i'$ given by Eq.~\eqref{eq:general} and applying this inequality along with \eqref{eq:piineq1} and~\eqref{eq:piineq2}, we find for $0\le\alpha\le1$ that
\begin{align}
f(\pi_i') &= \alpha \sum_j w_{ij} \log \pi_i'
  + (1-\alpha) \sum_j w_{ij} \log {\pi_i'\over\pi_i'+\pi_j}
  - \sum_j (\alpha w_{ij}+w_{ji}) \log (\pi_i'+\pi_j) \nonumber\\
  &\ge \alpha \sum_j w_{ij} \biggl[ \log \pi_i + {\pi_i'-\pi_i\over\pi_i'} 
   \biggr]
  + (1-\alpha) \sum_j w_{ij} \biggl[ \log {\pi_i\over\pi_i+\pi_j}
  + {(\pi_i'-\pi_i)/\pi_i'\over(\pi_i+\pi_j)/\pi_j} \biggr] \nonumber\\
  &\qquad{} - \sum_j (\alpha w_{ij}+w_{ji}) \biggl[ \log (\pi_i+\pi_j)
  + {\pi_i'-\pi_i\over\pi_i+\pi_j} \biggr] \nonumber\\
  &= f(\pi_i) + (\pi_i'-\pi_i)
     \biggl[ {1\over\pi_i'} \sum_j w_{ij} {\alpha\pi_i+\pi_j\over\pi_i+\pi_j}
     - \sum_j {\alpha w_{ij}+w_{ji}\over\pi_i+\pi_j} \biggr] \nonumber\\
  &= f(\pi_i),
\label{eq:genproof}
\end{align}
where we have employed Eq.~\eqref{eq:fgeneral} again, the term in square brackets in the penultimate line vanishes because of Eq.~\eqref{eq:general}, and the exact equality applies if and only if $\pi_i'=\pi_i$.  Thus $f(\pi_i)$ always increases upon application of~\eqref{eq:general} unless~$\pi_i'=\pi_i$.

The rest of the proof follows the same lines as in Section~\ref{sec:convergence} and hence convergence to the global likelihood maximum is established.  As a corollary, this also provides an alternative proof of the convergence of Zermelo's algorithm (the case $\alpha=1$) which is significantly simpler than the original proof given by \citet{Zermelo29} or the later proof by \citet{Ford57}.

The method of proof used here does not extend to the case of $\alpha>1$, because $1-\alpha$ becomes negative and the inequality in~\eqref{eq:genproof} no longer follows from~\eqref{eq:piineq1}.  It is still possible to prove convergence for $\alpha>1$ but the proof is more involved.  See Appendix~A for details.

Numerical measurements, some of which are presented in Section~\ref{sec:results}, indicate that convergence of the algorithms of this section becomes monotonically slower with increasing~$\alpha$, so that the main algorithm presented in this paper, Eq.~\eqref{eq:new}, which corresponds to the smallest allowed value of $\alpha=0$, is the fastest, and it is on this case that we concentrate in the remainder of the paper.  Some formal results on rates of convergence as a function of $\alpha$ are presented in Appendix~B.

\section{Prior on the strength parameters}
\label{sec:prior}
Equation~\eqref{eq:new} provides a complete algorithm for fitting the Bradley-Terry model.  In practice, however, pure maximum-likelihood fits such as this can be problematic for this model.  In particular, as mentioned in Section~\ref{sec:convergence}, a likelihood maximum exists only if the network of interactions is strongly connected.  If this condition is not met then the score parameters~$s_i$ will diverge and the algorithm of Eq.~\eqref{eq:new}---and indeed all maximum-likelihood methods for this model---will fail.

The root cause of this problem is that the maximum-likelihood fit effectively assumes a uniform (improper) prior on the~$s_i$, which places all but a vanishing fraction of its weight on arbitrarily large values and, when coupled with a network that is not strongly connected, causes divergences.  An effective solution is to impose a better-behaved prior on $s_i$ and then compute a maximum a posteriori (MAP) estimate of the scores instead of a maximum-likelihood estimate (MLE).  A range of priors have been proposed for this purpose~\citep{DS73,CD12,Whelan17} but arguably the most natural is a logistic prior.  Recall from Section~\ref{sec:bt} that the probability~$p_1$ of a player with strength~$\pi$ winning against the average player is $p_1 = \pi/(\pi+1)$.  In the absence of any evidence to the contrary, we assume this probability to be uniformly distributed between zero and one so that $P(p_1)=1$, a least informative or maximum-entropy prior.  Then the prior on the scores~$s = \log \pi$~is
\begin{equation}
P(s) = P(p_1) {dp_1\over ds} = {dp_1\over d\pi} {d\pi\over ds}
  = {\pi\over(\pi+1)^2}
  = {1\over(e^s+1)(e^{-s}+1)},
\label{eq:logistic}
\end{equation}
which is the logistic distribution.  Combining this result with Eq.~\eqref{eq:likelihood} we then get a posterior probability on the scores that is given, up to a multiplicative constant, by
\begin{align}
P(s|W) &\propto \prod_{ij} \biggl( {e^{s_i}\over e^{s_i}+e^{s_j}}
               \biggr)^{w_{ij}}
               \prod_i {1\over(e^{s_i}+1)(e^{-s_i}+1)} \nonumber\\
  &= \prod_{ij} \biggl( {\pi_i\over\pi_i+\pi_j}
               \biggr)^{w_{ij}}
               \prod_i {\pi_i\over(\pi_i+1)^2}.
\label{eq:map}
\end{align}
Maximizing this posterior probability instead of the likelihood regularizes the values of the scores, preventing them from diverging.  It also removes the invariance under multiplication of the $\pi_i$ by a constant and hence eliminates the need to normalize them.

The iterative algorithm of Eq.~\eqref{eq:new} can be generalized straightforwardly to this MAP estimate.  As observed by \citet{Whelan17}, the prior for individual~$i$ can be thought of~as
\begin{equation}
{\pi_i\over(1+\pi_i)^2}
  = {\pi_i\over\pi_i+1} \times {1\over\pi_i+1},
\end{equation}
which is precisely the probability that $i$ plays two games against the average player (who has~$\pi=1$) and wins one of them and loses the other.  Thus Eq.~\eqref{eq:map} can be thought of as the likelihood of a Bradley-Terry model in which we have added two fictitious games for each player, one won and one lost, and we can maximize this likelihood (and hence the posterior of Eq.~\eqref{eq:map}) using the same algorithm as before, merely adding these extra fictitious games to the data.  This also means that our proof of convergence generalizes to the MAP case and that the network of interactions is now strongly connected, so the probability maximum always exists.

Alternatively, and perhaps more conveniently, we can derive an explicit algorithm for the MAP case by differentiating Eq.~\eqref{eq:map} with respect to $\pi_i$ for any~$i$, which leads to the iteration
\begin{equation}
\pi_i' = {1/(\pi_i+1) + \sum_j w_{ij} \pi_j/(\pi_i+\pi_j)
             \over 1/(\pi_i+1) + \sum_j w_{ji}/(\pi_i+\pi_j)}.
\label{eq:newmap}
\end{equation}
This is the generalization of Eq.~\eqref{eq:new} to the MAP case. It is completely equivalent to adding the fictitious games and has the same guaranteed convergence.  One can also add the same prior to the traditional Zermelo algorithm of Eq.~\eqref{eq:zermelo}, which gives
\begin{equation}
\pi_i' = {1 + \sum_j w_{ij}\over2/(\pi_i+1)
             + \sum_j (w_{ij}+w_{ji})/(\pi_i+\pi_j)}.
\label{eq:zermmap}
\end{equation}
In Section~\ref{sec:results} we present the results of numerical experiments on the rate of convergence both of these MAP estimators and of the MLEs, using Eqs.~\eqref{eq:zermelo}, \eqref{eq:new}, \eqref{eq:newmap}, and~\eqref{eq:zermmap}.

\section{Ties}
\label{sec:ties}
Ties or draws can occur in certain types of competition, such as chess and soccer.  There are a number of ways to generalize ranking calculations to include ties.  The simplest is just to consider a tied game to be half of a win for each of the players.  This approach is used for instance in the Elo chess rating system and can be trivially incorporated into our calculations by modifying the values~$w_{ij}$.  A more sophisticated approach, however, incorporates the probability of a tie into the model itself.  There is more than one way to do this \citep{RK67,Davidson70}.  Here we employ the modification of the Bradley-Terry model proposed by \citet{Davidson70}.  One again defines strengths~$\pi_i$ for each player and the probabilities of a win~$p_{ij}$ and a tie~$q_{ij}$ between players~$i$ and~$j$ are
\begin{equation}
p_{ij} = {\pi_i\over\pi_i+\pi_j+2\nu\sqrt{\pi_i\pi_j}}, \qquad
q_{ij} = {2\nu\sqrt{\pi_i\pi_j}\over\pi_i+\pi_j+2\nu\sqrt{\pi_i\pi_j}},
\label{eq:ties}
\end{equation}
where $\nu>0$ is a parameter which controls the overall frequency of ties and which we will estimate by maximum likelihood along with the strengths.  Note that when $\pi_i=\pi_j$ we have $q_{ij} = \nu/(1+\nu)$ and hence $\nu = q_{ij}/(1-q_{ij})$, so $\nu$ can be interpreted as the odds of a tie between evenly matched players.

The form~\eqref{eq:ties} satisfies the obvious requirements that $p_{ij}+p_{ji}+q_{ij}=1$ and $q_{ij}=q_{ji}$, and also has the intuitive property that the probability of a tie is greatest when the players are evenly matched and vanishes as $\pi_i$ and~$\pi_j$ become arbitrarily far apart.  As with the standard Bradley-Terry model, the probabilities~$p_{ij}$ and~$q_{ij}$ are invariant under multiplication of all~$\pi_i$ by a constant, and again we remove this ambiguity by normalizing them so that $\prod_i \pi_i = 1$.

\citet{Davidson70} proposed an iterative algorithm for computing maximum-likelihood estimates of the strengths and the parameter~$\nu$ within this model.  Defining $w_{ij}$ as before to be the number of times $i$ beats~$j$ and $t_{ij}=t_{ji}$ to be the number of ties, we can write the likelihood of a set of observations $W=[w_{ij}]$, $T=[t_{ij}]$ as
\begin{align}
P(W,T|\pi,\nu) &= \prod_{ij} p_{ij}^{w_{ij}} \prod_{i<j} q_{ij}^{t_{ij}} \nonumber\\
  &= \prod_{ij} \biggl( {\pi_i\over\pi_i+\pi_j+2\nu\sqrt{\pi_i\pi_j}}
     \biggr)^{w_{ij}} \prod_{i<j} \biggl(
     {2\nu\sqrt{\pi_i\pi_j}\over\pi_i+\pi_j+2\nu\sqrt{\pi_i\pi_j}}
     \biggr)^{t_{ij}},
\label{eq:llties}
\end{align}
and the corresponding log-likelihood is
\begin{align}
\log P(W,T|\pi,\nu)
  &= \sum_{ij} \bigl( w_{ij} + \tfrac12 t_{ij} \bigr) \log \pi_i
     + \tfrac12\log2\nu \sum_{ij} t_{ij} \nonumber\\
  &\hspace{4em}{} - \sum_{ij} \bigl( w_{ij} + \tfrac12 t_{ij} \bigr) 
   \log \bigl( \pi_i+\pi_j+2\nu\sqrt{\pi_i\pi_j} \bigr).
\end{align}
The combination $w_{ij}+\frac12 t_{ij}$ comes up repeatedly in the analysis so, following Davidson, we define the convenient shorthand~$a_{ij} = w_{ij}+\frac12 t_{ij}$ and
\begin{equation}
\log P(W,T|\pi,\nu)
   = \sum_{ij} a_{ij} \log \pi_i
     + \tfrac12\log2\nu \sum_{ij} t_{ij}
     - \sum_{ij} a_{ij} \log \bigl( \pi_i+\pi_j+2\nu\sqrt{\pi_i\pi_j} \bigr).
\label{eq:liketies}
\end{equation}
Differentiating with respect to~$\pi_i$ and setting the result to zero gives
\begin{align}
{1\over\pi_i} \sum_j a_{ij}
  = \sum_j \bigl( a_{ij}+a_{ji} \bigr) {1 + \nu\sqrt{\pi_j/\pi_i}\over
          \pi_i+\pi_j+2\nu\sqrt{\pi_i\pi_j}}.
\label{eq:pities}
\end{align}
This equation has no general closed-form solution for~$\pi_i$ but Davidson proposed solving it by the obvious iteration
\begin{equation}
\pi_i' = {\sum_j a_{ij} \over
  \sum_j \bigl( a_{ij}+a_{ji} \bigr) \biggl( {\displaystyle 1
  + \nu\sqrt{\vphantom{\pi_i}\smash{\pi_j/\pi_i}}\over
    \displaystyle\pi_i+\pi_j+2\nu\sqrt{\pi_i\pi_j}} \biggr) }.
\label{eq:davidsonpi}
\end{equation}
We can also calculate a maximum-likelihood estimate of the parameter~$\nu$ by differentiating~\eqref{eq:liketies} with respect to~$\nu$ to get
\begin{equation}
{1\over2\nu} \sum_{ij} t_{ij}
  = \sum_{ij} a_{ij} {2\sqrt{\pi_i\pi_j}\over\pi_i+\pi_j+2\nu\sqrt{\pi_i\pi_j}},
\label{eq:nuties}
\end{equation}
which is again solved by iteration:
\begin{equation}
\nu' = {\frac12 \sum_{ij} t_{ij}\over
  \sum_{ij} a_{ij} \biggl(
  {\displaystyle2\sqrt{\pi_i\pi_j}\over
   \displaystyle\pi_i+\pi_j+2\nu\sqrt{\pi_i\pi_j}} \biggr) }.
\label{eq:davidsonnu}
\end{equation}
Davidson used asynchronous updates in which one applies Eq.~\eqref{eq:davidsonpi} to each~$\pi_i$ in turn, then applies~\eqref{eq:davidsonnu} once to update~$\nu$, then repeats until convergence is achieved.  This is a natural generalization of Zermelo's algorithm, Eq.~\eqref{eq:zermelo}, to situations where ties are allowed, and it includes Zermelo's algorithm as the special case when $\nu=0$ and $t_{ij}=0$.  Davidson proved that the procedure always converges to the global likelihood maximum (when the maximum exists), but once again convergence can be slow in practice.  Here we propose an alternative algorithm which generalizes Eq.~\eqref{eq:new} and is substantially faster.

Equation~\eqref{eq:pities} can be rearranged in the form
\begin{equation}
{1\over\pi_i} \sum_j a_{ij} {\pi_j + \nu\sqrt{\pi_i\pi_j}\over
          \pi_i+\pi_j+2\nu\sqrt{\pi_i\pi_j}}
  = \sum_j a_{ji} {1 + \nu\sqrt{\pi_j/\pi_i}\over
          \pi_i+\pi_j+2\nu\sqrt{\pi_i\pi_j}},
\end{equation}
which can be solved by iterating the equation
\begin{equation}
\pi_i' = {\sum_j a_{ij} \biggl(
  {\displaystyle\pi_j+\nu\sqrt{\pi_i\pi_j}\over
   \displaystyle\pi_i+\pi_j+2\nu\sqrt{\pi_i\pi_j}} \biggr) \over
  \sum_j a_{ji} \biggl( {\displaystyle1 +
  \nu\sqrt{\vphantom{\pi_i}\smash{\pi_j/\pi_i}}\over
  \displaystyle\pi_i+\pi_j+2\nu\sqrt{\pi_i\pi_j}} \biggr) }.
\label{eq:newties}
\end{equation}
Similarly, writing $a_{ij} = w_{ij} + \frac12 t_{ij}$, Eq.~\eqref{eq:nuties} can be rearranged in the form
\begin{equation}
{1\over2\nu} \sum_{ij} t_{ij} {\pi_i+\pi_j\over\pi_i+\pi_j+2\nu\sqrt{\pi_i\pi_j}}
  = \sum_{ij} w_{ij} {2\sqrt{\pi_i\pi_j}\over\pi_i+\pi_j+2\nu\sqrt{\pi_i\pi_j}},
\end{equation}
which can be solved by iterating
\begin{equation}
\nu' = { \frac12 \sum_{ij} t_{ij}
  \biggl( {\displaystyle\pi_i+\pi_j\over
  \displaystyle\pi_i+\pi_j+2\nu\sqrt{\pi_i\pi_j}} \biggr) \over
  \sum_{ij} w_{ij} \biggl( {\displaystyle2\sqrt{\pi_i\pi_j}\over
  \displaystyle\pi_i+\pi_j+2\nu\sqrt{\pi_i\pi_j}} \biggr) }.
\label{eq:nu}
\end{equation}

Equations~\eqref{eq:newties} and~\eqref{eq:nu} are the appropriate generalization of~\eqref{eq:new} to the case with ties and they include~\eqref{eq:new} as the special case when $\nu=0$ and $t_{ij}=0$.  Again we recommend applying the equations asynchronously: one cycle of the algorithm involves updating each~$\pi_i$ in turn using Eq.~\eqref{eq:newties} then applying Eq.~\eqref{eq:nu} once to update~$\nu$, and repeating until convergence is achieved.  As we show in Section~\ref{sec:results}, this procedure converges significantly faster than Davidson's algorithm.

The proof that Eqs.~\eqref{eq:newties} and~\eqref{eq:nu} do in fact converge to the likelihood maximum follows similar lines to that for the case without ties but the algebra is tedious so we omit it here.  The interested reader can find it in Appendix~C.

\section{Results}
\label{sec:results}
The iterations~\eqref{eq:new} and~\eqref{eq:newmap} (for the case without ties) and~\eqref{eq:newties} and~\eqref{eq:nu} (with ties) converge significantly faster in typical applications than the traditional algorithm of Zermelo or its extension for the case where ties are allowed.  In this section we illustrate the convergence rates with a selection of example applications to both real and synthetic data.

\subsection{Computer-generated data}
\label{sec:synthetic}
As our first example, we apply our algorithms to a collection of random computer-generated data sets.  In these calculations we generated synthetic test data with $N=1000$ players and $M=50\,000$ games, for an average of 100 games per player.  The players for each game are chosen uniformly at random (with replacement) and the winners of the games are chosen using the Bradley-Terry model itself: scores~$s_i$ for each player are drawn from a logistic distribution $P(s)=1/[(e^s+1)(e^{-s}+1)]$ and then the winner of each game is chosen at random according to the probability~$p_{ij}$ of Eq.~\eqref{eq:btwin}.  In cases where the resulting network of interactions is not strongly connected, games are discarded and redrawn until a strongly connected network is achieved, to ensure that a likelihood maximum does exist as discussed in Section~\ref{sec:convergence}.

\begin{figure}[t]
\begin{center}
\includegraphics[width=\textwidth]{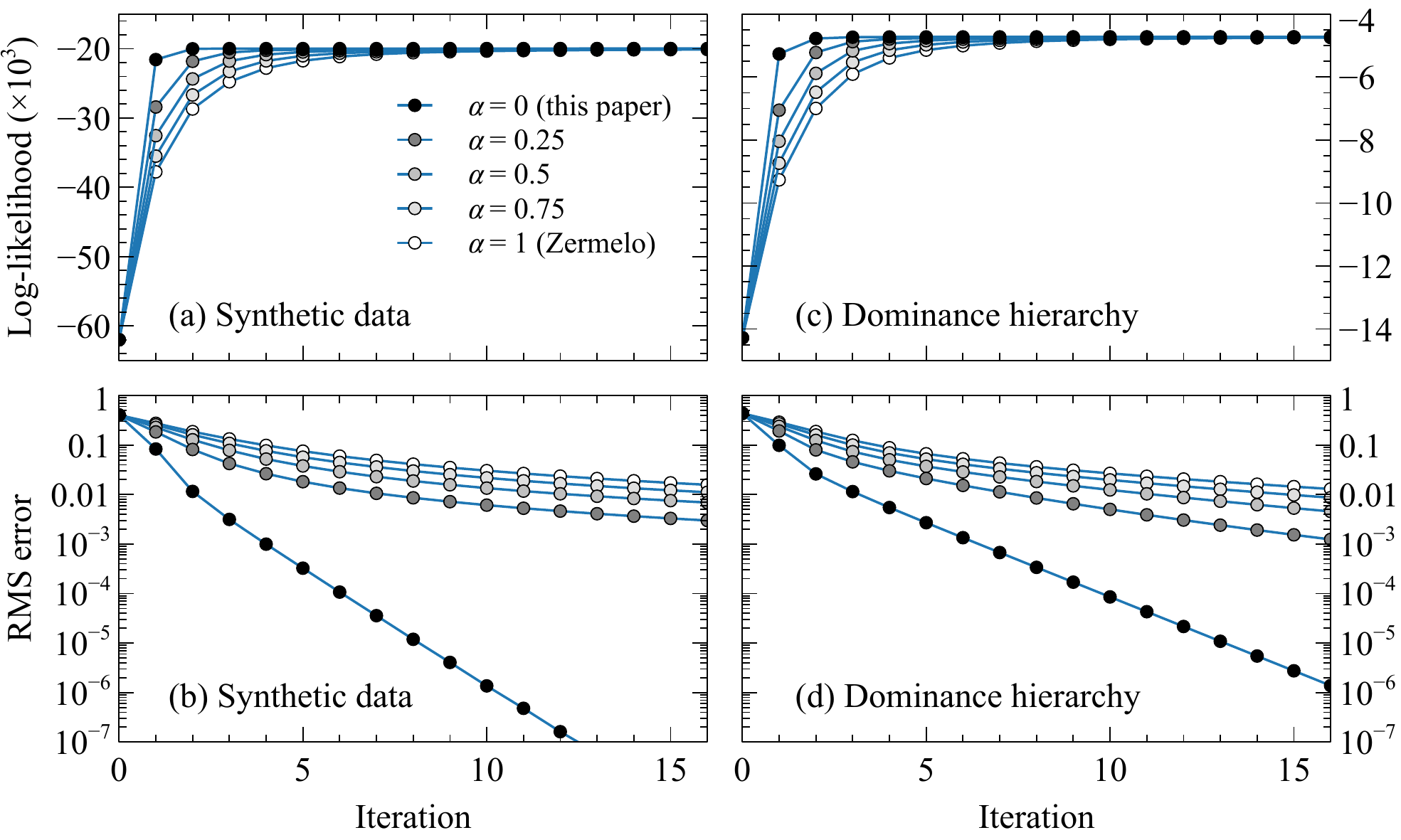}
\end{center}
\caption{Convergence of the iterative algorithms studied here.  (a)~The log-likelihood, Eq.~\eqref{eq:loglike}, for a synthetic network of 1000 players and 50\,000 games.  The plot shows the value on successive iterations for the algorithm of this paper, Zermelo's algorithm, and a selection of algorithms from the family defined in Section~\ref{sec:general}.  (b)~The root-mean-square (RMS) deviation from the final maximum-likelihood solution for the same synthetic data set.  (c)~The log-likelihood for the dominance hierarchy of vervet monkeys described in Section~\ref{sec:real}.  (d)~The RMS deviation for the same dominance hierarchy data set.}
\label{fig:convergence}
\end{figure}

Figure~\ref{fig:convergence}a shows the results of applying both Zermelo's algorithm and the algorithm of this paper to one such synthetic network.  Initial values of $\pi_i$ were chosen randomly such that $s_i$ is drawn from the logistic distribution $1/[(e^s+1)(e^{-s}+1)]$.  (Other methods of choosing the initial values have been proposed and may improve performance in some cases \citep{Dykstra56}, but we avoid these here to separate effects of the different algorithms from effects of the initial values.)  Figure~\ref{fig:convergence}a shows how the log-likelihood, Eq.~\eqref{eq:loglike}, converges to its final value on successive iterations of each algorithm.  As we can see, the algorithm of this paper (top curve, solid points) comes close to the final value of the log-likelihood after only two iterations, while Zermelo's algorithm (bottom curve, open points) takes significantly longer.  The other curves in the figure (gray points) show values for various algorithms in the family defined in Section~\ref{sec:general}, parametrized by the quantity~$\alpha$ as in Eq.~\eqref{eq:general}, and it appears that convergence becomes monotonically slower with increasing~$\alpha$, as mentioned previously in Section~\ref{sec:general}.

Arguably, however, Fig.~\ref{fig:convergence}a fails to truly show how much faster the algorithm of this paper is.  Figure~\ref{fig:convergence}b gives an alternative view.  In this plot we show the root-mean-square (RMS) difference (averaged over all players) between the current estimate of the probability $p_1=\pi_i/(\pi_i+1)$ of beating the average player and the final converged value.  We choose this quantity because the probability of beating the average player is typically more uniformly distributed than the~$\pi_i$ themselves and hence the average over players is better behaved, minimizing effects of fluctuations and dependency on outliers.  As the figure shows this quantity converges enormously faster for our algorithm than for Zermelo's algorithm, being already more than a thousand times better than the value for Zermelo's algorithm after less than ten iterations.  Indeed there is a significant gulf in the speed of convergence between our algorithm and all of the others shown in the figure: for all the nonzero values of $\alpha$ convergence is much slower than it is for $\alpha=0$.

To make these comparisons more quantitative, we have conducted extensive further tests using synthetic data of the type described here.  In these tests we first iterate Eq.~\eqref{eq:new} until it converges to high precision, effectively solving for the maximum-likelihood solution at or close to the limits of numerical accuracy.  Then, using either the Zermelo algorithm or our new algorithm, we measure the number of iterations needed to converge to this final solution within a prescribed level of accuracy.  Specifically we require that the probability $\pi_i/(\pi_i+1)$ of beating the average player converge to within $10^{-6}$ from its final value for all~$i$.  This criterion is more stringent---and arguably more realistic---than criteria based on convergence of the value of the log-likelihood \citep{VYZ19}.

For tests of the algorithm of Section~\ref{sec:prior} for MAP estimates with a logistic prior the same procedure was used to generate data and we compare the convergence of the new algorithm of Eq.~\eqref{eq:newmap} against the generalized Zermelo algorithm of Eq.~\eqref{eq:zermmap}.  For the case with ties the procedure was similar, but wins, losses, and ties were generated according to the probabilities $p_{ij}$ and $q_{ij}$ of Eq.~\eqref{eq:ties} with $\nu=\frac12$ and we compare the convergence rate of Eqs.~\eqref{eq:newties} and~\eqref{eq:nu} against Eqs.~\eqref{eq:davidsonpi} and~\eqref{eq:davidsonnu} with an initial value of $\nu=1$ in all cases.

All tests were averaged over 100 randomly generated data sets and the results are reported in Table~\ref{tab:results}.  As these results show, the algorithm of this paper is much faster than Zermelo's algorithm.  For the standard maximum likelihood estimate (the most common application) the difference is spectacular: the new algorithm is over a hundred times faster.  Where Zermelo's algorithm takes an average of more than 1200 iterations to converge, the new algorithm takes just~12.  For the MAP estimates the difference in running time is less extreme but still large---the new algorithm is over eight times faster than Zermelo's algorithm---while for the case with ties the new algorithm is an impressive 42 times faster.

\begin{table}[t]
\begin{center}
\begin{tabular}{ll|lrr|ccc}
& & & & & \multicolumn{3}{c}{Iterations to reach convergence} \\
& & Data              & $N$      & $M$      & Zermelo  & This paper & Speed-up \\
\hline
\begin{rotate}{90}
\hbox{\hspace{-62pt}Without ties}
\end{rotate} &
\begin{rotate}{90}
\hbox{\hspace{-46pt}MLE}
\end{rotate}
  & Synthetic         & 1000     & 50\,000  & $1270\pm470$    & $12\pm2$    & $\times104$ \\
& & Wolves            & 15       & 10\,382  & $2410\pm10$     & $145\pm1$   & $\times17$  \\
& & Vervet monkeys    & 63       & 11\,621  & $232\pm8$       & $19\pm1$    & $\times12$  \\
& & American football & 32       & 35\,741  & $49\pm3$        & $15\pm1$    & $\times3.4$ \\
& & Political figures & 67       & 76\,632  & $54\pm3$        & $8\pm5$     & $\times7.1$ \\
& & Photographs       & 9097     & 247\,531 & $911\pm4$       & $22\pm0$    & $\times41$ \\
\hline
\begin{rotate}{90}
\hbox{\hspace{-38pt}Without}
\end{rotate} &
\begin{rotate}{90}
\hbox{\hspace{-40pt}ties MAP}
\end{rotate}
  & Synthetic         & 1000     & 50\,000  & $1560\pm40$      & $185\pm18$    & $\times8.5$ \\
& & Wolves            & 15       & 10\,382  & $49\,200\pm1700$ & $2200\pm110$  & $\times22$ \\
& & American football & 32       & 35\,741  & $19\,200\pm1500$ & $6000\pm600$  & $\times3.3$ \\
& & Photographs       & 9097     & 247\,531 & $1186\pm3$       & $82\pm10$     & $\times14$ \\
\hline
&
\begin{rotate}{90}
\hbox{\hspace{-40pt}With ties}
\end{rotate}
  & Synthetic         & 1000     & 50\,000  & $1130\pm760$     & $27\pm8$      & $\times42$  \\
& & Soccer            & 177      & 898      & $1650\pm16$      & $421\pm5$     & $\times3.9$ \\
& & School students   & 2155     & 8970     & $2770\pm10$      & $613\pm1$     & $\times4.5$ \\
& & Chess             & 14\,852  & 623\,727 & $1750\pm90$      & $162\pm9$     & $\times11$ \\
\end{tabular}
\end{center}
\caption{The number of iterations required for the algorithms discussed in this paper to converge in applications to real and synthetic (computer-generated) data.  Results are averaged over 100 runs and rounded to three figures.  $N$~is the number of individuals or teams being ranked, $M$~is the total number of interactions among all individuals, and the figures following ``$\pm$'' are standard deviations about the mean, giving an indication of the amount of variation in the results.  ``Speed-up'' is the average factor by which the method of this paper improves upon the traditional Zermelo algorithm, or its generalizations for the MAP case and the case where ties are allowed.}
\label{tab:results}
\end{table}

\subsection{Real-world data}
\label{sec:real}
In this section we present example applications to several real-world data sets and show that our algorithm also offers significant speed improvements in these settings.  The data sets we study are as follows.

\paragraph*{Wolves:} A typical animal dominance hierarchy data set describing observations of subordinate behaviors among members of a family of 15 captive wolves in Arnhem, Netherlands as reported by \citet{VW87}.

\paragraph*{Vervet monkeys:} A larger dominance hierarchy data set describing observations of agonistic interactions of various kinds among 63 wild vervet monkeys in the Samara Private Game Reserve in South Africa, as reported by \citet{VBHB20}.  The original data set had 66 monkeys, but three were removed in order to ensure that the network of interactions was strongly connected, as discussed in Section~\ref{sec:convergence}.

\paragraph*{American football:} As an example of an application to sports competition, this data set describes professional American football games played in the US National Football League during a single season.  Unlike association football, American football proceeds by a series of discrete plays in which the team currently in possession of the ball attempts to advance it up the field.  This data set consists of individual plays in all games between the 32 teams in the league during the 2016 regular season, as compiled by \citet{YVH19}.  Only passing plays, running plays, punts, sacks, and field goals were used in the analysis.  Other plays such as kickoffs and conversions were excluded.  The team in possession of the ball is considered to have won a play if either (a)~they score points on the play or (b)~they advance the ball and retain possession; otherwise the other team wins the play.

\paragraph*{Political figures:} The results of an online paired comparison survey conducted by the Washington Post newspaper in 2010, in which readers were presented with pairs of prominent political figures and asked to judge which had had the worse week in politics.  The data were made available on the survey platform allourideas.org.

\paragraph*{Photographs:} Results from the IMDB-Wiki-SbS study of \citet{PU21}, a paired comparison study that asked participants to judge people's age from photographs.   Participants were presented with 247\,531 pairs of faces drawn from a pool of 9097 photo\-graphs and asked to judge which of the people depicted was older.  In principle, a ranking of the results should then be able to order the people from (apparent) oldest to youngest.  A small number of images were excluded from the data set for our calculations to ensure a strongly connected network.

\paragraph*{Soccer:} Wins, losses, and draws in 898 men's international association football matches between 177 different countries during the year 2011.  Data from Mart J\"urisoo at kaggle.com/martj42.  The original network of matches was not strongly connected, so the data analyzed here represent only the largest strongly connected component of the network.

\paragraph*{School students:} These data describe declared friendships among 2155 students in a large US high school and its feeder middle school, from the National Longitudinal Study of Adolescent Health (the ``Add Health'' study, \citealt{Addhealth}).  If student~$i$ states that they are friends with student~$j$ but $j$ does not reciprocate (something that occurs often in these data) we consider it a win for~$j$; if $i$ and~$j$ both state they are friends we consider it a tie.  Although in principle friendships are not competitive, there is evidence that friendship patterns among school students do describe a clear hierarchy because students tend to claim friendship with others who have higher social status than themselves \citep{HK88}.  Thus ranking calculations applied to data like these can be used to infer social status \citep{BN13}.  Treating a reciprocated friendship as a tie is arguably more correct than treating it as two separate wins: reciprocated friendships clearly violate the assumption of independence in the Bradley-Terry model without ties, since the two wins never go in the same direction, but there is no equivalent violation for the model with ties.  The same approach could also be applied to other social networks that show similar reciprocity properties.  The network of friendships for this data set was not strongly connected, so the data analyzed here represent only the largest strongly connected component of the network.

\paragraph*{Chess:} Wins, losses, and draws in chess matches between 14\,852 expert players on the online chess server lichess.com during the month of July 2016.  For a match to be included, both players must have had Elo ratings of 2000 or higher at the time of the match.  A small number of players were removed to ensure the network of matches was strongly connected.  The data are from lichess.com via kaggle.com/arevel.  With over 600\,000 matches, this is the largest data set considered here.

\vspace{3ex} Figures~\ref{fig:convergence}c and~\ref{fig:convergence}d show an example of the convergence of the log-likelihood and RMS error during a single run using the vervet monkey data.  The behavior is similar to that for the synthetic data in Figs.~\ref{fig:convergence}a and~\ref{fig:convergence}b: the log-likelihood converges most rapidly for the algorithm of this paper and significantly more slowly for Zermelo's algorithm.  Other algorithms from the family defined in Section~\ref{sec:general} fall between the two, and convergence appears to become monotonically slower as the parameter~$\alpha$ of Eq.~\eqref{eq:general} increases.  The RMS error shown in Fig.~\ref{fig:convergence}d once again shows very rapid convergence for the algorithm of this paper.  All the other algorithms are substantially slower by this measure.

Complete results on time to convergence for the various data sets are presented in Table~\ref{tab:results}.  The methodology for these calculations was the same as for the synthetic data: the parameters were first converged to high precision, then the results used to estimate the time to convergence in a second run of the calculation.  Each calculation was replicated 100 times with random initial conditions in which the $s_i$ were drawn from a logistic distribution as previously.

The overall picture for these runs is again similar to that for the synthetic data.  In all cases the method of this paper outpaces the traditional Zermelo algorithm.  For instance, for maximum-likelihood estimates in cases without ties the new algorithm is 17 times faster on the smallest example, the dominance hierarchy of wolves, while on the largest example, the photographs, it is a remarkable 41 times faster.  The smallest difference is for the American football data set, for which the new algorithm is 3.4 times faster than Zermelo's algorithm.  For MAP estimates the numbers are similar: the new algorithm is a factor of 22, 14, and 3.3 times faster respectively on these three data sets.

As with the synthetic data, the speed difference on the tests with ties is less dramatic though still substantial, with the new algorithm being about 4 to 11 times faster.  Convergence was also somewhat slower overall for both algorithms in the case with ties, although this may have more to do with the fact that these data sets are sparser (which tends to slow convergence) than with the presence of ties.  Notice that convergence of our algorithm is very fast for the synthetic data with ties, which is relatively dense.

These effects can make a substantial difference to running times in practice.  For the dominance hierarchy of wolves, for instance, a single run of Zermelo's algorithm (implemented in the Python programming language on an up-to-date but otherwise unremarkable personal computer \textit{circa}~2022) converges to the maximum-likelihood solution in a running time of about 1 minute.  The algorithm of this paper, by contrast, takes 3 seconds.  For the more demanding photograph data set, Zermelo's algorithm takes over 8 minutes; the method of this paper takes just 11 seconds.  For larger applications still, such as to web data or online social networks, the difference could become very significant.

All the results of this section are numerical.  Ideally we would like to be able to prove formally that the algorithms presented in this paper converge faster than Zermelo's algorithm.  At present we do not have such a proof but we can show certain results.  As demonstrated in Appendix~B, we can prove that within the one-parameter family of algorithms defined in Section~\ref{sec:general}, all those for $\alpha>1$ converge slower than Zermelo's algorithm (the case $\alpha=1$), which means these are not normally of interest.  We can also prove that convergence becomes monotonically faster with decreasing~$\alpha$ down to some point $\alpha<1$, meaning that there provably exist algorithms that are faster than Zermelo's algorithm.  In general, however, the proof does not extend to $\alpha=0$ (the algorithm of this paper), so for the moment the finding that convergence is fastest for $\alpha=0$ is a numerical one only.

\section{Conclusions}
\label{sec:conclusions}
We have presented an alternative to the classic algorithm of Zermelo for computing rankings from pairwise comparisons using fits to the Bradley-Terry model, with or without ties allowed.  Like Zermelo's algorithm, the method presented is a simple iterative scheme.  We have proved that the iteration always converges to the global maximum of the likelihood and given numerical evidence that it does so faster---typically many times faster---than Zermelo's algorithm.  Given that it is also simple to implement we know of no reason not to favor the algorithm presented here over Zermelo's algorithm.

\section*{Acknowledgments}
This work was funded in part by the US National Science Foundation under grant DMS--2005899.  All empirical data used in this paper are previously published and freely available online.

\appendix
\section*{Appendix A: Proof of convergence for $\alpha>1$}
As discussed in Section~\ref{sec:general}, Zermelo's algorithm and the algorithm of this paper are both special cases of a larger one-parameter family of algorithms given by the iteration of
\begin{equation}
\pi_i' = {\sum_j w_{ij} (\alpha\pi_i+\pi_j)/(\pi_i+\pi_j)\over
            \sum_j (\alpha w_{ij}+w_{ji})/(\pi_i+\pi_j)}
\end{equation}
for any $\alpha\ge0$.  For $0\le\alpha\le1$ the convergence of this iteration to the likelihood maximum can be proved straightforwardly as described in Section~\ref{sec:general}.  For $\alpha>1$ the same method of proof does not work because $1-\alpha$ becomes negative and the inequality in~\eqref{eq:genproof} no longer follows from~\eqref{eq:piineq1}.  It is still possible to prove convergence but the method of proof is somewhat different, as we now describe.

From~\eqref{eq:ineq2} we have for any $x,y,c>0$
\begin{align}
\log(x+c) &\ge \log(y+c) - {y+c\over x+c} + 1 = \log(y+c) + {x-y\over x+c}
  \nonumber\\
  &= \log(y+c) + {x-y\over x} - {(x-y)/x\over(y+c)/c}
     + {c(x-y)^2\over x(x+c)(y+c)} \nonumber\\
  &\ge \log(y+c) + {x-y\over x} - {(x-y)/x\over(y+c)/c},
\end{align}
which is equivalent to
\begin{equation}
\log x - \log y - {x-y\over x} \ge \log {x\over x+c} - \log {y\over y+c}
  - {(x-y)/x\over(y+c)/c},
\end{equation}
with the exact equality applying if and only if $x=y$.  Noting that the left-hand side of this inequality is always positive by~\eqref{eq:ineq2}, for any $\alpha>1$ we then have
\begin{equation}
\log x - \log y - {x-y\over x} \ge {\alpha-1\over\alpha} \biggl[
  \log {x\over x+c} - \log {y\over y+c} - {(x-y)/x\over(y+c)/c} \biggr],
\end{equation}
which can be rearranged to read
\begin{equation}
\alpha \log x + (1-\alpha) \log {x\over x+c}
  \ge \alpha \biggl[ \log y + {x-y\over x} \biggr]
  + (1-\alpha) \biggl[ \log {y\over y+c} + {(x-y)/x\over(y+c)/c} \biggr].
\end{equation}
Now setting $x=\pi_i'$, $y=\pi_i$, and $c=\pi_j$, multiplying by the positive quantities~$w_{ij}$, and summing, we have
\begin{align}
\alpha \sum_j w_{ij} \log \pi_i'
  + (1-\alpha) \sum_j w_{ij} \log {\pi_i'\over\pi_i'+\pi_j}
  &\ge \alpha \sum_j w_{ij} \biggl[ \log \pi_i + {\pi_i'-\pi_i\over\pi_i'} 
   \biggr] \nonumber\\
  &{} + (1-\alpha) \sum_j w_{ij} \biggl[ \log {\pi_i\over\pi_i+\pi_j}
  + {(\pi_i'-\pi_i)/\pi_i'\over(\pi_i+\pi_j)/\pi_j} \biggr],
\end{align}
where the exact equality applies if and only if $\pi_i'=\pi_i$.  In combination with~\eqref{eq:piineq2}, this is now sufficient to establish the inequality in~\eqref{eq:genproof} once again, and hence convergence of the algorithm for $\alpha>1$ is assured.

\section*{Appendix B: Rate of convergence}
The numerical results of Section~\ref{sec:results} show markedly faster convergence for the algorithms of this paper than for the standard Zermelo algorithm.  As discussed at the end of Section~\ref{sec:results}, we do not at present have a proof that convergence is faster, but it is possible to prove that some algorithms within the family defined in Section~\ref{sec:general} converge faster than Zermelo's algorithm.

As observed in Fig.~\ref{fig:convergence}, the iterative algorithms of this paper show exponential convergence, which is expected---in general all iterations of the form $x' = f(x)$ converge exponentially, if they converge at all, except in certain special cases that do not apply here.  For the family of algorithms in Section~\ref{sec:general} the rate of convergence for any given value of the parameter~$\alpha$ can be quantified by the factor~$\lambda_i(\alpha)$ by which the distance between the current estimate of $\pi_i$ and the final maximum-likelihood estimate (MLE)~$\hat{\pi}_i$ decreases when $\pi_i$ is updated, as $\pi_i$ approaches~$\hat{\pi}_i$.  Thus
\begin{equation}
\lambda_i(\alpha)
  = \lim_{\pi\to\hat{\pi}} {\pi_i'-\hat{\pi}_i\over\pi_i-\hat{\pi}_i}
  = \biggl( {\partial\pi_i'\over\partial\pi_i} \biggr)_{\!\hat{\pi}},
\label{eq:lambda}
\end{equation}
where the subscript~$\hat{\pi}$ indicates that the derivative is evaluated at the MLE.  For instance, for Zermelo's algorithm (the case $\alpha=1$), applying Eq.~\eqref{eq:tired} we have
\begin{equation}
\lambda_i(1) = {\sum_j w_{ij} \sum_j (w_{ij}+w_{ji})/(\hat{\pi}_i+\hat{\pi}_j)^2
  \over\bigl[ \sum_i (w_{ij}+w_{ji})/(\hat{\pi}_i+\hat{\pi}_j) \bigr]^2}
  = {1\over\sum_j w_{ij}}
  \sum_j (w_{ij}+w_{ji}) \biggl( {\hat{\pi}_i\over\hat{\pi}_i+\hat{\pi}_j}
  \biggr)^2,
\end{equation}
where we have employed~\eqref{eq:tired} again to simplify the expression and made use of the fact that $\pi_i'=\pi_i=\hat{\pi}_i$ at the MLE.  Assuming once again that the network of interactions represented by~$w_{ij}$ is strongly connected, the $w_{ij}$ are strictly positive for all~$i,j$.  As shown by \citet{Ford57}, this implies that the $\hat{\pi}_i$ are strictly positive and finite, which means in turn that the value of $\lambda_i(1)$ is strictly positive.  For other~$\alpha$, however, the value of $\lambda_i(\alpha)$ can be negative (meaning that convergence to the MLE is oscillatory).  This observation will be important in a moment.

The factor by which the RMS error of Fig.~\ref{fig:convergence} decreases over a complete round of updates depends asymptotically on the slowest decaying~$\pi_i$ and is given by
\begin{equation}
\lambda_\textrm{max}(\alpha) = \max_i|\lambda_i(\alpha)|,
\end{equation}
where we take the absolute value to allow for the possibility of negative~$\lambda_i$.  An algorithm with given~$\alpha$ asymptotically converges faster than Zermelo's algorithm if $\lambda_\textrm{max}(\alpha)<\lambda_\textrm{max}(1)$.  Here we demonstrate that this is the case for at least some values of~$\alpha$.

We consider how the value~$\lambda_i(\alpha)$ changes with~$\alpha$ and compute the derivative
\begin{equation}
{\partial\lambda_i\over\partial\alpha}
  = {\partial\over\partial\alpha} \biggl( {\partial\pi_i'\over\partial\pi_i}
    \biggr)_{\!\hat{\pi}}
  = \biggl( {\partial^2\pi_i'\over\partial\pi_i\partial\alpha}
    \biggr)_{\!\hat{\pi}}.
\end{equation}
From Eq.~\eqref{eq:general} we have
\begin{align}
{\partial\pi_i'\over\partial\alpha}
  &= \biggl[ \sum_j {\alpha w_{ij} + w_{ji}\over\pi_i+\pi_j}
     \sum_j w_{ij} {\pi_i\over\pi_i+\pi_j}
     - \sum_j w_{ij} {\alpha\pi_i+\pi_j\over\pi_i+\pi_j}
     \sum_j {w_{ij}\over\pi_i+\pi_j} \biggr] \Bigg/
     \biggl[ \sum_j {\alpha w_{ij} + w_{ji}\over\pi_i+\pi_j} \biggr]^2
     \nonumber\\
  &= {\sum_j w_{ij}/(\pi_i+\pi_j) \over
     \sum_j (\alpha w_{ij}+w_{ji})/(\pi_i+\pi_j)} (\pi_i - \pi_i'),
\end{align}
where we have used~\eqref{eq:general} again in the second line.  Differentiating with respect to~$\pi_i$, setting~$\pi_i=\hat{\pi}_i$ for all~$i$, and noting again that $\pi_i'=\pi_i$ at the MLE, we find that
\begin{equation}
{\partial\lambda_i\over\partial\alpha}
  = \biggl( {\partial^2\pi_i'\over\partial\pi_i\partial\alpha}
    \biggr)_{\!\hat{\pi}}
  = {\sum_j w_{ij}/(\hat{\pi}_i+\hat{\pi}_j) \over
     \sum_j (\alpha w_{ij}+w_{ji})/(\hat{\pi}_i+\hat{\pi}_j)}
     \bigl[ 1 - \lambda_i(\alpha) \bigr],
\label{eq:dlambda}
\end{equation}
where we have used Eq.~\eqref{eq:lambda}.

The fact that the iteration of Eq.~\eqref{eq:general} converges to the MLE for all~$\alpha\ge0$ implies that $\lambda_i(\alpha)$ must be strictly less than~1 for all~$i$ and hence~\eqref{eq:dlambda} is strictly positive, since $w_{ij}$ and $\hat{\pi}_i$ are strictly positive.  At the same time it is also finite, and hence $\lambda_i(\alpha)$ is increasing in~$\alpha$ and continuous for all $\alpha\ge0$.

This now establishes some useful results.  First, it implies that $\lambda_i(\alpha)>\lambda_i(1)$ for all~$i$ when $\alpha>1$ (and also that $\lambda_i(\alpha)$ is positive in this regime).  Thus, if the largest value of $\lambda_i(1)$ occurs for $i=\mu$, then
\begin{equation}
\lambda_\textrm{max}(\alpha) \ge \lambda_\mu(\alpha) > \lambda_\mu(1)
  = \lambda_\textrm{max}(1).
\end{equation}
Hence all algorithms with $\alpha>1$ converge slower than Zermelo's algorithm.  For this reason these algorithms are not normally of practical interest.

Second, we also have $\lambda_i(\alpha)<\lambda_i(1)$ for all~$i$ when $0\le\alpha<1$.  Unfortunately, this is not sufficient to establish that $\lambda_\textrm{max}(\alpha)<\lambda_\textrm{max}(1)$ in this regime (and hence that these algorithms converge faster than Zermelo's algorithm) because, as mentioned above, it is not guaranteed that $\lambda_i(\alpha)$ is positive.  The value of $\lambda_i(\alpha)$ for $\alpha<1$ can---and in practice often does---become negative.  This means that $|\lambda_i(\alpha)|$ could be larger than~$\lambda_i(1)$ and indeed it is straightforward to find cases where this occurs.

On the other hand, we can prove that there exist \emph{some} algorithms that are faster than Zermelo's.  Given that $\lambda_i(\alpha)$ is continuous and increasing in~$\alpha$, its value must diminish smoothly and monotonically from $\alpha=1$ all the way down to $\alpha=0$.  Thus, given that $\lambda_i(1)$ is strictly positive, one of two things must happen: either $\lambda_i(\alpha)$ never reaches the line $\lambda_i(\alpha)=-\lambda_i(1)$, in which case $|\lambda_i(0)|<\lambda_i(1)$, or it \emph{does} reach this line, in which case there exists some $c_i<1$ such that $\lambda_i(c_i)=-\lambda_i(1)$.  In this case, by continuity, $|\lambda_i(\alpha)|<\lambda_i(1)$ in the non-vanishing interval $c_i<\alpha<1$.

Now we repeat the same argument for all~$i$ and define $c = \max_i c_i$, or $c=0$ if $|\lambda_i(0)|<\lambda_i(1)$ for all~$i$, and then for all $i$ we have $|\lambda_i(\alpha)| < \lambda_i(1)$ in the non-vanishing interval $c<\alpha<1$.  Now choose any $\alpha$ in this interval and suppose the largest value of $|\lambda_i(\alpha)|$ occurs for $i=\nu$.  Then at this $\alpha$ we have
\begin{equation}
\lambda_\textrm{max}(\alpha) = |\lambda_\nu(\alpha)| < \lambda_\nu(1)
  \le \lambda_\textrm{max}(1).
\end{equation}
Hence all algorithms with $c<\alpha<1$ converge faster than Zermelo's algorithm.  Algorithms with $0\le\alpha\le c$ may also converge faster than Zermelo's algorithm---and the numerical evidence suggests that they do---but this cannot be proved using the present approach.

\section*{Appendix C: Proof of convergence for the model with ties}
For the case where ties are allowed, the proof that iteration of Eqs.~\eqref{eq:newties} and~\eqref{eq:nu} converges to the maximum of the log-likelihood~\eqref{eq:liketies} follows similar lines to that for the case without ties.  \citet{Davidson70} proved that the likelihood has only a single stationary point with respect to its parameters, corresponding to the global likelihood maximum, provided the $\pi_i$ are normalized and the network of interactions is strongly connected (with a tie counting as an edge in both directions between the relevant pair of players).  Since any fixed point of Eqs.~\eqref{eq:newties} and~\eqref{eq:nu} corresponds to a stationary point of the likelihood, this implies that if our iteration converges to a fixed point at all then that point is the global maximum.  To prove that we converge to a fixed point it suffices to show that the log-likelihood always increases upon application of either Eq.~\eqref{eq:newties} or Eq.~\eqref{eq:nu}, unless a fixed point has been reached.

The terms in the log-likelihood of Eq.~\eqref{eq:liketies} that depend on~$\pi_i$ can be written in the form
\begin{equation}
f(\pi_i) = \sum_j a_{ij} \log {\pi_i\over\pi_i+\pi_j+2\nu\sqrt{\pi_i\pi_j}}
   - \sum_j a_{ji} \log \bigl( \pi_i+\pi_j+2\nu\sqrt{\pi_i\pi_j} \bigr),
\label{eq:fties}
\end{equation}
where $a_{ij} = w_{ij} + \frac12 t_{ij}$ as previously.  Applying the inequalities~\eqref{eq:ineq1} and~\eqref{eq:ineq2}, we have for any~$\pi_i$ and~$\pi_i'$
\begin{align}
\log {\pi_i'\over\pi_i'+\pi_j+2\nu\sqrt{\pi_i'\pi_j}} &\ge \log {\pi_i\over\pi_i+\pi_j+2\nu\sqrt{\pi_i\pi_j}}
  - {\pi_i/(\pi_i+\pi_j+2\nu\sqrt{\pi_i\pi_j})\over\pi_i'/(\pi_i'+\pi_j+2\nu\sqrt{\pi_i'\pi_j})} + 1 \nonumber\\
  &= \log {\pi_i\over\pi_i+\pi_j+2\nu\sqrt{\pi_i\pi_j}}
     + \biggl( {\sqrt{\pi_i'}-\sqrt{\pi_i}\over\sqrt{\pi_i'}} \biggr) {2\nu\sqrt{\pi_i\pi_j}\over\pi_i+\pi_j+2\nu\sqrt{\pi_i\pi_j}} \nonumber\\
  &\hspace{6em}{} + \biggl( {\pi_i'-\pi_i\over\pi_i'} \biggr) {\pi_j\over\pi_i+\pi_j+2\nu\sqrt{\pi_i\pi_j}}
\end{align}
and
\begin{align}
\log \bigl( {\textstyle\pi_i'+\pi_j+2\nu\sqrt{\pi_i'\pi_j}} \bigr) &\le \log \bigl( {\textstyle\pi_i+\pi_j+2\nu\sqrt{\pi_i\pi_j}} \bigr)
  + {\pi_i'+\pi_j+2\nu\sqrt{\pi_i'\pi_j}\over\pi_i+\pi_j+2\nu\sqrt{\pi_i\pi_j}} - 1 \nonumber\\
  &= \log \bigl( {\textstyle\pi_i+\pi_j+2\nu\sqrt{\pi_i\pi_j}} \bigr)
     + \biggl( {\sqrt{\pi_i'}-\sqrt{\pi_i}\over\sqrt{\pi_i}} \biggr) {2\nu\sqrt{\pi_i\pi_j}\over\pi_i+\pi_j+2\nu\sqrt{\pi_i\pi_j}} \nonumber\\
  &\hspace{6em}{} + {\pi_i'-\pi_i\over\pi_i+\pi_j+2\nu\sqrt{\pi_i\pi_j}}.
\end{align}
Evaluating Eq.~\eqref{eq:fties} at the point~$\pi_i'$ defined by Eq.~\eqref{eq:newties} and applying these two inequalities, we have
\begin{align}
f(\pi_i') &= \sum_j a_{ij} \log {\pi_i'\over\pi_i'+\pi_j+2\nu\sqrt{\pi_i'\pi_j}} - \sum_j a_{ji} \log \bigl( \pi_i'+\pi_j+2\nu\textstyle\sqrt{\pi_i'\pi_j} \bigr) \nonumber\\
  &\ge \sum_j a_{ij} \biggl[  \log {\pi_i\over\pi_i+\pi_j+2\nu\sqrt{\pi_i\pi_j}}
     + \biggl( {\sqrt{\pi_i'}-\sqrt{\pi_i}\over\sqrt{\pi_i'}} \biggr) {2\nu\sqrt{\pi_i\pi_j}\over\pi_i+\pi_j+2\nu\sqrt{\pi_i\pi_j}} \nonumber\\
  &\hspace{20em}{} + \biggl( {\pi_i'-\pi_i\over\pi_i'} \biggr) {\pi_j\over\pi_i+\pi_j+2\nu\sqrt{\pi_i\pi_j}} \biggr] \nonumber\\
  &\qquad{} - \sum_j a_{ji} \biggl[ \log \bigl( {\textstyle\pi_i+\pi_j+2\nu\sqrt{\pi_i\pi_j}} \bigr)
     + \biggl( {\sqrt{\pi_i'}-\sqrt{\pi_i}\over\sqrt{\pi_i}} \biggr) {2\nu\sqrt{\pi_i\pi_j}\over\pi_i+\pi_j+2\nu\sqrt{\pi_i\pi_j}} \nonumber\\
  &\hspace{20em}{} + {\pi_i'-\pi_i\over\pi_i+\pi_j+2\nu\sqrt{\pi_i\pi_j}} \biggr] \nonumber\\
  &= f(\pi_i) + \sum_{ij} a_{ij} \biggl[ \biggl( {\sqrt{\pi_i'}-\sqrt{\pi_i}\over\sqrt{\pi_i'}} \biggr) {2\nu\sqrt{\pi_i\pi_j}\over\pi_i+\pi_j+2\nu\sqrt{\pi_i\pi_j}}
     - \biggl( {\pi_i'-\pi_i\over\pi_i'} \biggr) {\nu\sqrt{\pi_i\pi_j}\over\pi_i+\pi_j+2\nu\sqrt{\pi_i\pi_j}} \biggr] \nonumber\\
  &\hspace{3.65em}{} - \sum_j a_{ji} \biggl[ \biggl( {\sqrt{\pi_i'}-\sqrt{\pi_i}\over\sqrt{\pi_i}} \biggr) {2\nu\sqrt{\pi_i\pi_j}\over\pi_i+\pi_j+2\nu\sqrt{\pi_i\pi_j}}
     - \biggl( {\pi_i'-\pi_i\over\pi_i} \biggr) {\nu\sqrt{\pi_i\pi_j}\over\pi_i+\pi_j+2\nu\sqrt{\pi_i\pi_j}} \biggr] \nonumber\\
  &= f(\pi_i) + {(\sqrt{\pi_i'}-\sqrt{\pi_i})^2\over\pi_i'} \sum_j a_{ij} {\nu\sqrt{\pi_i\pi_j}\over\pi_i+\pi_j+2\nu\sqrt{\pi_i\pi_j}} \nonumber\\
  &\hspace{8em}{} + {(\sqrt{\pi_i'}-\sqrt{\pi_i})^2\over\pi_i} \sum_j a_{ji} {\nu\sqrt{\pi_i\pi_j}\over\pi_i+\pi_j+2\nu\sqrt{\pi_i\pi_j}} \nonumber\\
  &\ge f(\pi_i),
\end{align}
where we have used Eq.~\eqref{eq:newties} and the exact equality applies if and only if $\pi_i'=\pi_i$.  Hence $f(\pi_i)$ always increases upon application of Eq.~\eqref{eq:newties} unless $\pi_i'=\pi_i$, and so therefore does the log-likelihood as well.

The same is also true of the update~\eqref{eq:nu} for the parameter~$\nu$.  The terms in the log-likelihood that depend on~$\nu$ can be written
\begin{equation}
g(\nu)
  = \tfrac12 \sum_{ij} t_{ij} \log {\nu\over\pi_i+\pi_j+2\nu\sqrt{\pi_i\pi_j}}
    - \sum_{ij} w_{ij} \log(\pi_i+\pi_j+2\nu\sqrt{\pi_i\pi_j}).
\label{eq:g}
\end{equation}
For any $\nu,\nu'$ the inequalities~\eqref{eq:ineq1} and~\eqref{eq:ineq2} imply that
\begin{align}
\log {\nu'\over\pi_i+\pi_j+2\nu'\sqrt{\pi_i\pi_j}}
  &\ge \log {\nu\over\pi_i+\pi_j+2\nu\sqrt{\pi_i\pi_j}}
      - {\nu/(\pi_i+\pi_j+2\nu\sqrt{\pi_i\pi_j})\over
         \nu'/(\pi_i+\pi_j+2\nu'\sqrt{\pi_i\pi_j})} + 1 \nonumber\\
  &= \log {\nu\over\pi_i+\pi_j+2\nu\sqrt{\pi_i\pi_j}}
     + \biggl( {\nu'-\nu\over\nu'} \biggr)
     {\pi_i+\pi_j\over\pi_i+\pi_j+2\nu\sqrt{\pi_i\pi_j}}, \\
\log(\pi_i+\pi_j+2\nu'\sqrt{\pi_i\pi_j})
  &\le \log(\pi_i+\pi_j+2\nu\sqrt{\pi_i\pi_j})
       + {\pi_i+\pi_j+2\nu'\sqrt{\pi_i\pi_j}\over
          \pi_i+\pi_j+2\nu\sqrt{\pi_i\pi_j}} - 1 \nonumber\\
  &= \log(\pi_i+\pi_j+2\nu\sqrt{\pi_i\pi_j})
       + (\nu'-\nu) {2\sqrt{\pi_i\pi_j}\over\pi_i+\pi_j+2\nu\sqrt{\pi_i\pi_j}}.
\end{align}
Evaluating~\eqref{eq:g} at the point~$\nu'$ given by Eq.~\eqref{eq:nu} and applying these two inequalities we get
\begin{align}
g(\nu')
 &= \tfrac12 \sum_{ij} t_{ij} \log {\nu'\over\pi_i+\pi_j+2\nu'\sqrt{\pi_i\pi_j}}
   - \sum_{ij} w_{ij} \log(\pi_i+\pi_j+2\nu'\sqrt{\pi_i\pi_j}) \nonumber\\
 &\ge \tfrac12 \sum_{ij} t_{ij} \biggl[
  \log {\nu\over\pi_i+\pi_j+2\nu\sqrt{\pi_i\pi_j}}
     + \biggl( {\nu'-\nu\over\nu'} \biggr)
      {\pi_i+\pi_j\over\pi_i+\pi_j+2\nu\sqrt{\pi_i\pi_j}} \biggr] \nonumber\\
 &\hspace{4em}{} - \sum_{ij} w_{ij}
     \biggl[ \log(\pi_i+\pi_j+2\nu\sqrt{\pi_i\pi_j})
       + (\nu'-\nu) {2\sqrt{\pi_i\pi_j}\over\pi_i+\pi_j+2\nu\sqrt{\pi_i\pi_j}}
     \biggr] \nonumber\\
 &= g(\nu) + (\nu'-\nu) \biggl[ {1\over2\nu'} \sum_{ij} t_{ij}
    {\pi_i+\pi_j\over\pi_i+\pi_j+2\nu\sqrt{\pi_i\pi_j}}
    - \sum_{ij} w_{ij}
      {2\sqrt{\pi_i\pi_j}\over\pi_i+\pi_j+2\nu\sqrt{\pi_i\pi_j}} \biggr],
    \nonumber\\
 &= g(\nu),
\end{align}
where the term in square brackets in the penultimate line vanishes because of Eq.~\eqref{eq:nu} and the exact equality applies if and only if $\nu'=\nu$.  Thus $g(\nu)$ always increases upon application of~\eqref{eq:nu} unless $\nu'=\nu$, and so therefore does the log-likelihood.

The remainder of the proof follows the same lines of argument as in Section~\ref{sec:convergence} and hence convergence of Eqs.~\eqref{eq:newties} and~\eqref{eq:nu} to the unique likelihood maximum is established.

\end{document}